\documentclass[10pt,twocolumn,letterpaper]{article}

\usepackage[pagenumbers]{cvpr} 

\usepackage{graphicx}
\usepackage{amsmath}
\usepackage{amssymb}
\usepackage{booktabs}
\usepackage{pifont}
\usepackage{multirow}
\usepackage{arydshln}
\usepackage[outercaption]{sidecap}    
\usepackage{color, colortbl}
\usepackage[title]{appendix}
\usepackage{enumitem}
\definecolor{LightGray}{rgb}{0.92,0.92,0.92}
\definecolor{Red}{rgb}{1.0, 0.13, 0.32}

\newcommand{\zyang}[1]{{\color[rgb]{0,0,1}{\tiny\textbf{ZY:}}{\normalsize\itshape#1}}}
\newcommand{\eat}[1]{}
\usepackage[pagebackref,breaklinks,colorlinks]{hyperref}

\usepackage[capitalize]{cleveref}
\crefname{section}{Sec.}{Secs.}
\Crefname{section}{Section}{Sections}
\Crefname{table}{Table}{Tables}
\crefname{table}{Tab.}{Tabs.}

\def\cvprPaperID{1446} 
\def\confName{CVPR}
\def\confYear{2022}

\begin{document}

\title{LaTr: Layout-Aware Transformer for Scene-Text VQA}

\author{
  Ali Furkan Biten\textsuperscript{1}\thanks{Authors contribute equally.} \qquad
  Ron Litman\textsuperscript{2}\footnotemark[1]\qquad
  Yusheng Xie\textsuperscript{2}\qquad
  Srikar Appalaraju\textsuperscript{2}\qquad
  R. Manmatha\textsuperscript{2}\qquad\\
  \textsuperscript{1}Computer Vision Center, UAB, Spain, \qquad\textsuperscript{2}Amazon Web Services\\
  {\tt\small abiten@cvc.uab.es}\qquad
  {\tt\small \{litmanr, yushx, srikara, manmatha\}@amazon.com}\\
}
\newcommand{\AlgoName}{LaTr }
\newcommand{\AlgoNameCombined}{LaTr$^\ddagger$ }
\newcommand{\AlgoNameCombinedNoSpace}{LaTr$^\ddagger$}
\newcommand{\AlgoNameNoSpace}{LaTr}
\newcommand{\OCRName}{Amazon-OCR }
\newcommand{\OCRNameNoSpace}{Amazon-OCR}
\newcommand{\TaskName}{STVQA }
\newcommand{\TaskNameNoSpace}{STVQA}
\maketitle

\begin{abstract}
We propose a novel multimodal architecture for Scene Text Visual Question Answering (\TaskNameNoSpace), named Layout-Aware Transformer (\AlgoNameNoSpace).
The task of \TaskName requires models to reason over different modalities. Thus, we first investigate the impact of each modality, and reveal 
the importance of the language module, especially when enriched with layout information. Accounting for this, we propose a single objective pre-training scheme that requires only text and spatial cues. We show that applying this pre-training scheme on scanned documents has certain advantages over using natural images, despite the domain gap. Scanned documents are easy to procure, text-dense and have a variety of layouts, helping the model learn various spatial cues (e.g. left-of, below etc.) by tying together language and layout information. 
Compared to existing approaches, our method performs vocabulary-free decoding and, as shown, generalizes well beyond the training vocabulary. We further demonstrate that \AlgoName improves robustness towards OCR errors, a common reason for failure cases in \TaskNameNoSpace.
In addition, by leveraging a vision transformer, we eliminate the need for an external object detector.
\AlgoName outperforms state-of-the-art \TaskName methods on multiple datasets. In particular, +7.6\% on TextVQA, +10.8\% on ST-VQA and +4.0\% on OCR-VQA (all absolute accuracy numbers).
\end{abstract}

\section{Introduction}

\begin{figure}[t]
    \centering
    \includegraphics[width=\columnwidth]{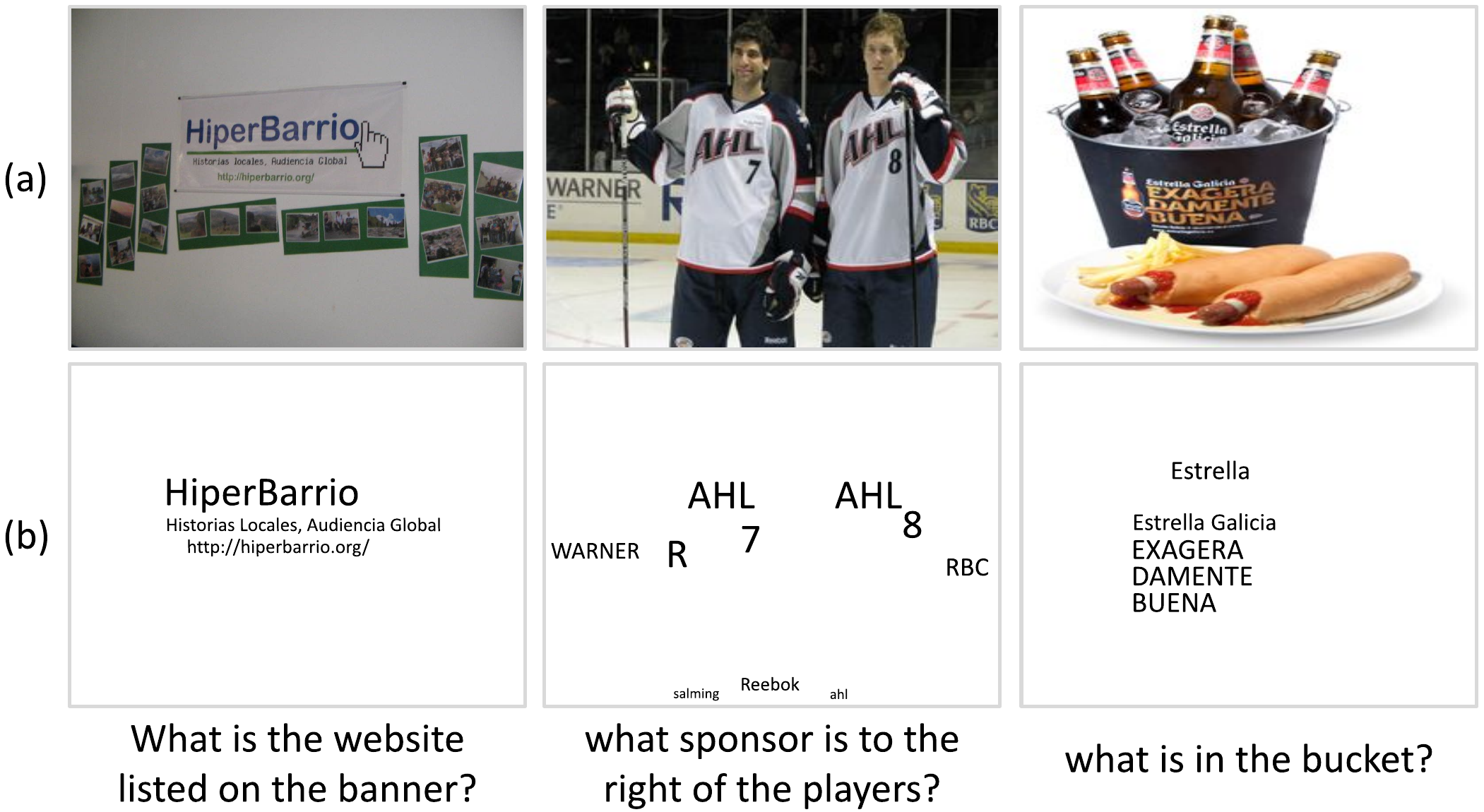}
    \caption{\textbf{The Role of Language and Layout in \TaskNameNoSpace.} (a) Representative samples from TextVQA. (b) We visualize the information extracted by the OCR system, showing that some questions only require text features, some require both text and layout information and only some need beyond that.
    Accounting for this, we propose a \textit{layout-aware} pre-training and architecture.}
    \label{fig:info_type_teaser}
    \vspace{-0.50cm}
\end{figure}

Scene-Text VQA (\TaskNameNoSpace) aims to answer questions by utilizing the scene text in the image.
It requires reasoning over rich semantic information conveyed by various modalities -- vision, language and scene text.
\cref{fig:info_type_teaser}~(a) depicts representative samples in \TaskNameNoSpace, showcasing a model's desired abilities, including; (1) a-priori information and world knowledge such as knowing what a website looks like (left image); and (2) the capability to use language, layout, and visual information (middle and right images).

In this work, we introduce Layout-Aware Transformer (\AlgoNameNoSpace), a multimodal encoder-decoder transformer based model for \TaskNameNoSpace. We begin by exploring how far language and layout information can take us in \TaskName.
In \cref{fig:info_type_teaser}~(b) we visualize the information extracted by the optical character recognition (OCR) system~\cite{baek2019wrong,litman2020scatter,aberdam2021sequence,fang2021read}, exhibiting three question categories: the first type can be answered with just the text tokens; the second type can be answered with text and layout information (\textit{right} vs \textit{left}); the third can only be answered by utilizing text, spatial and visual features all together. We quantitatively show that in the current datasets, most questions fall under the first two categories.
To methodologically show this, we first evaluate a zero-shot language model on \TaskName benchmarks, and then show that \AlgoName can already correctly answer over 50\% of the questions with only text tokens. Next, we show the performance gain achieved by enriching the language modality with layout information via our propose \textit{layout-aware} pre-training and architecture.

Recently, Yang et al.~\cite{yang2021tap} demonstrated the advantages in pre-training \TaskName models on natural images, proposing \textit{text-aware} pre-training (TAP) scheme, which is designed to foster multi-modal collaboration. 
Acquiring large quantities of natural images with text is challenging and hard to scale, as most natural images do not contain scene text. Even when they do, the amount of text is often sparse (previous statistics suggest a median of only 6 words per image~\cite{veit2016coco,yang2021tap}).
In addition, and more importantly, TAP did not account for the importance of aligning the layout information with the semantic representations when designing the pre-training objectives.

To counter these drawbacks, we propose \textit{layout-aware} pre-training based on a single objective using only text and spatial cues as input. Our pre-training forces the model to learn a joint representation which accounts for the interactions between text and layout information, benefiting the down-stream task of \TaskNameNoSpace. 
Despite the domain gap, we find that pre-training on documents has certain advantages over natural images. Scanned documents contain more text compared to natural images, therefore it is easier to scale the experiment and expose the model to more data. Words in documents are usually complete sentences, helping the model better learn semantics beyond a simple bag of words. Moreover, scanned documents provide varied layouts, leading to effective alignment between language and spatial features. 
Lastly, performing pre-training without visual features reduces computational complexity substantially.

Our model utilizes a vision transformer~\cite{dosovitskiy2020image} for extracting visual features, thus replacing the extensive need for an external object detector~\cite{hu2020iterative,kant2020spatially,yang2021tap}.
Moreover, in practice, current \TaskName models exploit a dataset-specific vocabulary with a pointer mechanism for decoding~\cite{gomez2021multimodal, hu2020iterative, kant2020spatially, yang2021tap}, creating an over-reliance on the fixed vocabulary and leaving no room for fixing OCR errors. Our model performs vocabulary-free decoding, does well even on answers out-of-vocabulary, and even overcomes OCR errors in some cases. 
\AlgoName outperforms the state-of-the-art \TaskName methods by large margins on multiple public benchmarks. 
To summarize, the key contributions of our work are:
\begin{enumerate}[nolistsep,leftmargin=*]
    \item We recognize the key role language and layout play in \TaskName and propose a \textit{layout-aware} pre-training and architecture to account for that.
    
    \item We pinpoint a new symbiosis between documents and \TaskName via pre-training. We show empirically that documents are beneficial for tying together language and layout information despite the huge domain gap.
    
    \item We show that existing methods perform poorly on out-of-vocabulary answers. \AlgoName does not require a vocabulary, does well even on answers that are not in the training vocabulary, and can even overcome OCR errors.

    \item We provide extensive experimentation and show the effectiveness of our method by advancing the state-of-the-art by +7.6\% on TextVQA and +10.8\% on ST-VQA and +4.0\% in OCR-VQA dataset.
\end{enumerate}

\section{Related Work}
\noindent
\textbf{Pre-training and Language Models.} The low cost of obtaining language text combined with the success of pre-training, language models~\cite{devlin2018bert, radford2018improving, liu2019roberta, raffel2019exploring} has shown remarkable success in machine translation, natural language understanding, question answering and more. Recently, numerous studies~\cite{lu2019vilbert,li2019visualbert,alberti2019fusion,li2020unicoder,tan2019lxmert,su2019vl,zhou2020unified,chen2019uniter,lu202012,li2020oscar,huang2020pixel,kim2021vilt,li2021align} showed the benefits of pre-training multi-modal architectures for vision and language tasks. Yang et al.~\cite{yang2021tap} demonstrated, for the first time, the effectiveness of pre-training in scene text VQA by using masked language modeling and image-text matching as pretext tasks.
In this paper, we show that tying together language and layout information via a simple \textit{layout-aware} pre-training scheme is beneficial for scene text VQA. Moreover, we perform pre-training over scanned documents and discover that, despite the domain gap, documents can be leveraged for task of \TaskNameNoSpace.

\begin{figure*}[htp!]
    \centering
    \includegraphics[width=0.95\textwidth]{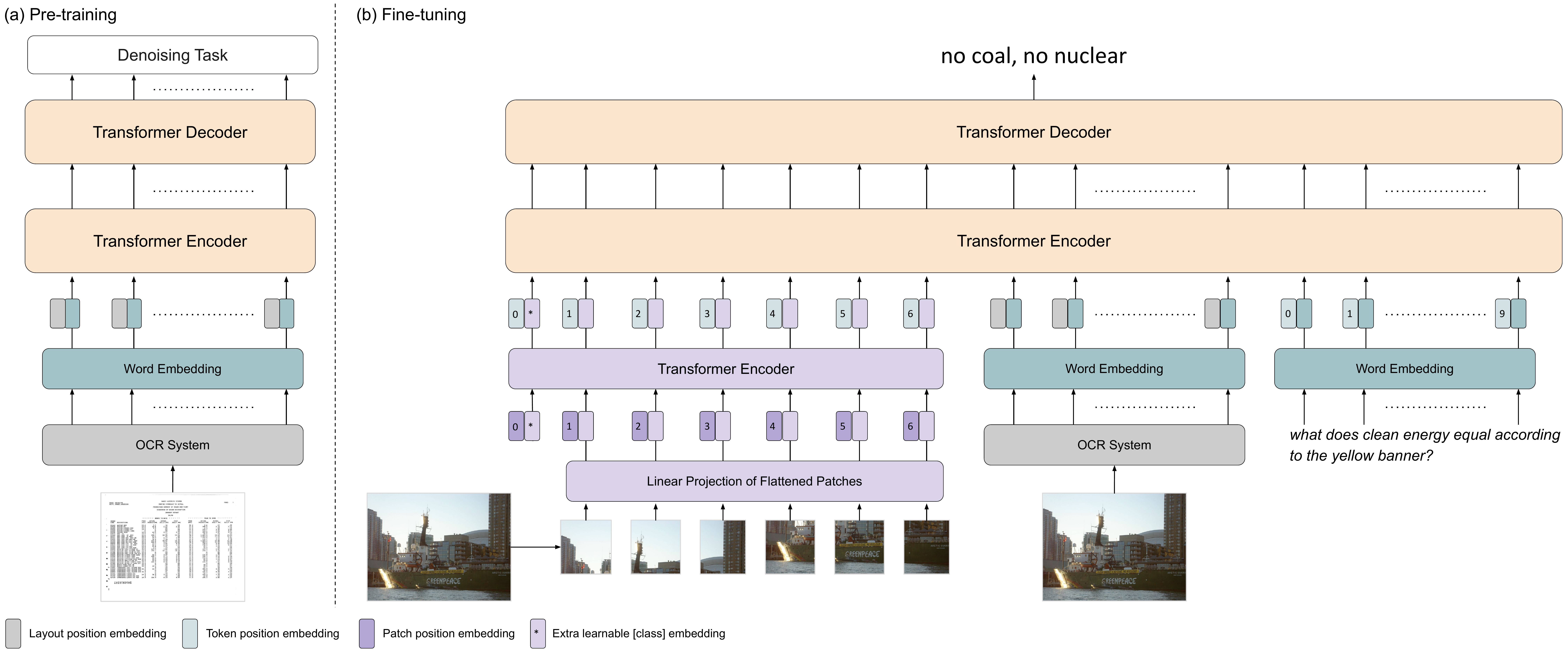}
    \caption{\textbf{An overview of \AlgoNameNoSpace.} (a) In pre-training, we only train the language modality with text and spatial cues to jointly model interactions between text and layout information. Pre-training is done on large amounts of documents. Documents are a text rich environment with a variety of layouts. (b) In fine-tuning, we add visual features from a ViT, thus eliminating the need for an external object detector.}
    \label{fig:main_figure}
    \vspace{-0.5cm}
\end{figure*}
\begin{figure}[t]
    \centering
    \includegraphics[width=\columnwidth]{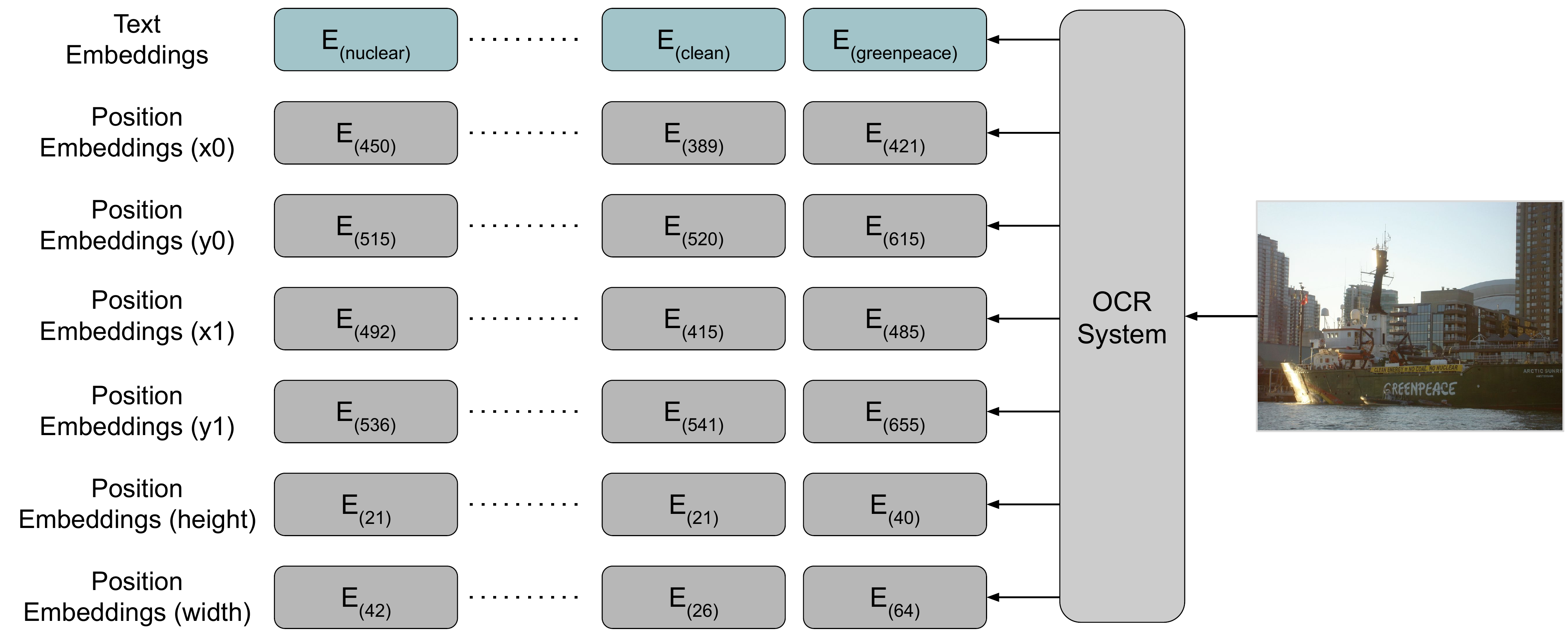}
    \caption{\textbf{Layout Position Embedding.}  2-D position embeddings representing the text layout in the image are leveraged to enrich the semantic representations.}
    \label{fig:layout_pos_emb}
    \vspace{-0.5cm}
\end{figure}

\noindent
\textbf{Vision-language tasks incorporating scene text.} Recently, integrating reading into the vision and language tasks has become imperative, especially in VQA and captioning where the models were known to be illiterate~\cite{biten2019icdar, singh2019towards}. 
Since the usage of text can be quite distinct in terms of the environment, several papers introduce new datasets for various contexts in which text appears; ST-VQA~\cite{biten2019scene}, TextVQA~\cite{singh2019towards} in natural images; OCR-VQA~\cite{mishra2019ocr} in book and movie covers;
DocVQA~\cite{mathew2021docvqa} in scanned documents; InfoVQA~\cite{mathew2021infographicvqa} in info-graphics. Moreover, STE-VQA~\cite{wang2020general} is proposed for multi-lingual VQA and TextCaps~\cite{sidorov2020textcaps} for captioning on natural images.
There are several papers published on scene text VQA. LoRRa~\cite{singh2019towards} extended Pythia~\cite{jiang2018pythia} with a pointer network~\cite{vinyals2015pointer} to select either from a fixed vocabulary or from OCR tokens. M4C~\cite{hu2020iterative} also used pointer networks but instead used multi-modal transformers~\cite{vaswani2017attention} to encode all modalities together. SA-M4C~\cite{kant2020spatially} build on top of M4C by providing supervision on self-attention weights. MM-GNN~\cite{gao2020multi} builds separate graphs for different modalities by utilizing graph neural networks~\cite{kipf2016semi}. Instead of having separate graphs for each modality, SMA~\cite{gao2020structured} introduces a single graph that encodes all modalities.~\cite{zhu2020simple} proposes to use an attention mechanism to fuse pairwise modalities.

\AlgoName enriches the language modality with layout information via pre-training to achieve state-of-the-art performance across multiple benchmarks. Our model is generative in nature and as such alleviates the problem of vocabulary reliance current methods suffer from. In addition, we will show that \AlgoName is more robust to OCR errors, one of the most common reasons for failure cases in \TaskNameNoSpace~\cite{hu2020iterative,yang2021tap}.

\section{Method}
In this section, we describe in detail our model architecture and our pre-training strategy, as seen in \cref{fig:main_figure}.  
\AlgoName consists of three main building blocks. First, a language model pre-trained on only text. Second, use of spatial embedding for OCR tokens bounding box in conjunction with further \textit{layout-aware} pre-training on documents, as depicted in \cref{fig:main_figure}~(a). Finally, a ViT architecture~\cite{dosovitskiy2020image} for obtaining visual features. We first explain each of the modules and then describe how all the modules come together as a whole.

\vspace{-5pt}

\paragraph{The Language Model}
We base our \AlgoName architecture on the encoder-decoder transformer architecture of \textit{Text-to-Text Transfer Transformer} (T5~\cite{raffel2019exploring}).
Apart from minor modifications, T5's architecture is roughly equivalent to the original transformer proposed by~\cite{vaswani2017attention}, which makes it easy to extend in various ways. In addition,  the vast amount of pre-training data used in the T5 pretraining makes it attractive for \TaskName as model initialization. In particular, \cite{raffel2019exploring} used Common Crawl publicly-available web archive to obtain a subset of 750 GB cleaned English text data, which they term Colossal Clean Crawled Corpus (C4). Pre-training on C4 is done with a de-noising task, which is a variant of masked-language modeling (MLM \cite{devlin2018bert}). We follow the implementation and use the weights from HuggingFace \cite{wolf2019huggingface}~\footnote{\url{https://huggingface.co/transformers/model\_doc/t5.html}}.

\paragraph{2-D Spatial Embedding}
Recent document understanding literature \cite{xu2020layoutlm, xu2020layoutlmv2, appalaraju2021docformer} prove the value of layout information when working with Transformers. The key idea is to associate and couple the 2-D positional information of the text with the language representation, \ie creating better alignment between the layout information and the semantic representation. 
Unlike words in a document, scene text in natural images may appear in arbitrary shapes and angles (e.g., as on a watch face). Therefore, we include the height and width of the text to indicate the reading order.

Formally, as seen in \cref{fig:layout_pos_emb}, given an OCR token $O_i$, the associated word bounding box may be defined by $(x^i_0, y^i_0, x^i_1, y^i_1, h^i, w^i)$, where $(x^i_0, y^i_0)$ corresponds to the position of the upper left corner of the bounding box, $(x^i_1, y^i_1)$ represents the position of the lower right corner, and $(h^i, w^i)$ represents the height and width with respect to the reading order. To embed bounding box information, we use a lookup table commonly used for continuous encoding one-hot representations (e.g. nn.Embedding in PyTorch).
Before we feed the word representation into the transformer encoder, we sum up all the representations together:
\begin{equation}
\begin{split}
        \mathcal{E}_i = \ &E_O(O_i) + E_{x}(x^i_0) + E_{y}(y^i_0) +\\ 
        & E_{x}(x^i_1) + E_{y}(y^i_1) + E_{w}(w^i) + E_h(h^i)
\end{split}
\label{eq:spatial-embed}
\end{equation}
where $\mathcal{E}_i$ is the encoded representation for an OCR token $O_i$ and $E_O, E_{x}, E_{y}, E_w, E_h$ are the learnable look-up tables.

\paragraph{Layout-Aware Pre-Training}
As T5 was trained on just text data, we perform further pre-training to effectively align the layout information (in form of the 2-D spatial embedding) and the semantic representations.
To the best of our knowledge, we are the first to propose pre-training on documents instead of natural images for the task of scene text VQA. The motivation for selecting documents is that they are a source of rich text environment in a variety of complex layouts. 
Inspired by \cite{raffel2019exploring}, we perform a \textit{layout-aware} de-noising pre-training task, which includes the 2-D spatial embedding, as seen in \cref{fig:main_figure}~(a).
This enables the use of weak data with no answer annotations in the pre-training stage.
Like the normal de-noising task, our \textit{layout-aware} de-noising task masks a span of tokens and forces the model to predict the masked spans. Unlike the normal de-noising task, we also give the model access to the rough location of the masked tokens, which encourages the model to fully utilize the layout information when completing this task.

More formally, let $\mathcal{O} = \{O_1, O_2, ..., O_n\}$ be the set of all OCR tokens (strings) and $\mathcal{B} = \{B_1, B_2 ..., B_n\}$ be the corresponding bounding box information, where $B_j = (x^j_0, y^j_0, x^j_1, y^j_1, w^j, h^j)$. Now, let $\mathcal{M}_l = \{j, j+1, ..., j+k\}$ be the $l^{th}$ mask span where $j$ is the starting index to mask such that $\max(M_l) < \min(M_{l+1})$. Then, $\{O_j, ..., O_{j+k}\}$ and $\{B_j, ..., B_{j+k}\}$ are replaced by $\Tilde{O_i}$ (a special indexed mask token) and $\Tilde{B_i}$ (the span's minimal containing bounding box) in the following manner:
\begin{equation}
\begin{split}
    \Tilde{O_i} = \ &\textless\text{extra\_id\_}l\textgreater,  \text{where} \ l\in \{0, ..., k-1\} \\
    \Tilde{B_i} = \ &(\min(\{x^i_0\}), \min(\{y^i_0\}), \\
    & \max(\{x^i_1\}), \max(\{y^i_1\})) \\
    & \text{where} \ {j\leq i \leq j+k}
\end{split}
\label{eq:masking}
\end{equation}
where the height and width of the masked tokens' bounding box are calculate with the coordinates of $\Tilde{B_i}$. 

Essentially, we have replaced a span of words tokens $\{O_j, ..., O_{j+k}\}$ and their corresponding bounding boxes $\{B_j, ..., B_{j+k}\}$ with a special token $\Tilde{O_i}$ and a corresponding "loose" bounding box. In other words, when we mask the span of words, we select the minimum of the top-left coordinates and the maximum of the bottom-right ones. The reasons are twofold. First, we do not want our model to know precise token boxes because that would reveal how many tokens are masked. Second, we choose not to mask the bounding boxes completely because then the model does not know where the text should appear in the document and cannot use the correct spatial context effectively. So, we prevent the model from taking shortcuts, but at the same time give it enough information to learn. 
The masked token $\Tilde{O_i}$ and its bounding box $\Tilde{B_i}$ are then embedded using \cref{eq:spatial-embed} like any other regular token.
We use cross-entropy loss to predict all the masked tokens' original text.

\begin{table*}
\normalsize
\begin{center}
\footnotesize
\bgroup
\def\arraystretch{1.1}
        \begin{tabular}{llllcc}
        \toprule
        \textbf{Method} & \textbf{OCR System} & \textbf{Pre-Training Data} & \textbf{Extra Finetune} & \textbf{Val Acc.} & \textbf{Test Acc.}\\
        \midrule
        M4C~\cite{hu2020iterative} & Rosetta-en & \ding{55}  &\ding{55}  & 39.40 & 39.01 \\
        SMA~\cite{gao2020structured} & Rosetta-en & \ding{55}  &\ding{55}  & 40.05 & 40.66 \\
        CRN~\cite{liu2020cascade} & Rosetta-en & \ding{55}  &\ding{55}  & 40.39 & 40.96 \\
        LaAP-Net~\cite{han2020finding} & Rosetta-en & \ding{55}  &\ding{55}  & 40.68 & 40.54 \\
        TAP~\cite{yang2021tap} & Rosetta-en & TextVQA &\ding{55}  & \eat{\zyang{44.06}}44.06 & - \\
        \rowcolor{LightGray} \AlgoName-Base  & Rosetta-en & \ding{55}  &\ding{55}  & 44.06  & - \\
        \rowcolor{LightGray} \AlgoName-Base  & Rosetta-en & IDL &\ding{55}  &  48.38 & - \\
        \midrule
        SA-M4C~\cite{kant2020spatially} & Google-OCR & \ding{55} & ST-VQA & 45.4 & 44.6 \\
        SMA~\cite{gao2020structured} & SBD-Trans OCR & \ding{55} & ST-VQA & - & 45.51 \\
        M4C~\cite{hu2020iterative,yang2021tap} &  Microsoft-OCR & \ding{55}  & ST-VQA & \eat{\zyang{45.22}} 45.22 & - \\
        TAP~\cite{yang2021tap} & Microsoft-OCR & TextVQA  &\ding{55}  & 49.91 & 49.71 \\
        TAP~\cite{yang2021tap} & Microsoft-OCR & TextVQA, ST-VQA & ST-VQA & 50.57 & 50.71 \\
        LOGOS~\cite{lu2021localize} & Microsoft-OCR & \ding{55} & ST-VQA & 51.53 & 51.08 \\
        TAP~\cite{yang2021tap} & Microsoft-OCR & TextVQA, ST-VQA, TextCaps, OCR-CC & ST-VQA & 54.71 & 53.97 \\
        
        M4C~\cite{hu2020iterative} &  \OCRNameNoSpace & \ding{55}  & \ding{55} & 47.84 & - \\
            
        \rowcolor{LightGray} \AlgoNameNoSpace-Base  & \OCRNameNoSpace & \ding{55}  & \ding{55}  & 52.29 & - \\

        \rowcolor{LightGray} \AlgoNameNoSpace-Base  & \OCRNameNoSpace & IDL  & \ding{55} & 58.03 & 58.86 \\
        \rowcolor{LightGray} \AlgoNameCombinedNoSpace-Base  & \OCRNameNoSpace & IDL & ST-VQA & 59.53 & 59.55 \\

        \rowcolor{LightGray} \AlgoNameNoSpace-Large  & \OCRNameNoSpace & IDL & \ding{55} & 59.76 & 59.24 \\
        \rowcolor{LightGray} \AlgoNameCombinedNoSpace-Large  & \OCRNameNoSpace & IDL & ST-VQA  & \textbf{61.05} & \textbf{61.60}\\
        \bottomrule
    \end{tabular}
\egroup
\tiny
\caption{\textbf{Results on the TextVQA dataset~\cite{singh2019towards}}. As commonly done, the top part of the table presents results in the constrained setting that only uses TextVQA
for training and Rosetta for OCR detection, while the bottom part is the unconstrained settings. \AlgoName advances the state-of-the-art performance, specifically by +6.43\% and +7.63\% on validation and test, respectively.}
\label{table:textvqa}
\vspace{-0.8cm}
\end{center}
\end{table*}
\vspace{-5pt}
\paragraph{Visual Features}
Most previous methods utilized an external pre-trained object detector~\cite{hu2020iterative, yang2021tap} for extracting objects labels, visual
object features and visual OCR features. In this work, we diverge from the literature and leverage a Vision Transformer (ViT)~\cite{dosovitskiy2020image}. The ViT is an image classification network which is pre-trained and fine-tuned on ImageNet~\cite{deng2009imagenet}. We utilize ViT in our architecture only in the fine-tuning stage, and we freeze all the layers except the last fully connected projection layer we add. Formally, an image $I$  having the dimension of ${H \times W \times C}$ is reshaped into 2D patches of size ${N \times (p^2\cdot C)}$, where $(H, W)$ is the height and width, $C$ is the number of channels, $(P, P)$ is the resolution of each image patch, and $N = HW/P^2$ is the final number of patches.
As depicted in \cref{fig:main_figure}~(b), we utilize a linear projection layer to map the flattened patches to $D$ dimensional space and feed them to the ViT.
We pass the full ViT output (containing $[class]$ token) sequence to a trainable linear projection layer and then feed it to the transformer encoder. Position embeddings are added to the patch embeddings to retain positional information. We denote the final visual output as $\mathcal{V} = \{V_0, ..., V_{N}\}$.

\vspace{-5pt}

\paragraph{\AlgoNameNoSpace}
So far, we explained the building blocks of our method, now we describe how we put it all together, as depicted in \cref{fig:main_figure}~(b). After pre-training the language modality of the model with layout information, we input all three modalities, namely; image, OCR information and question to the transformer encoder.
Let $\mathcal{V} = \{V_0, ..., V_{N}\}$ be a set of visual patch features such that $V_0$ is the $[class]$ embedding, $\mathcal{Q} = \{W_1, ..., W_m\}$ be the question tokenized into $W_i$ and $\mathcal{O} = \{O_1, O_2, ..., O_n\}$ be the OCR tokens. We embed the OCR tokens and questions using \cref{eq:spatial-embed} to obtain encoded OCR tokens $\mathcal{E}$ and encoded question features $\mathcal{E}^q$. For the 2-D spatial embedding of each $W_i$, we use fixed values ($x_0=y_0=0; x_1=y_1=1000$). Finally, we concatenate all the inputs $[\mathcal{V}; \mathcal{E}; \mathcal{E}^q]$ to feed to the multimodal transformer encoder-decoder architecture. Cross entropy loss is used to fine-tune our model. 

\section{Experiments}

\noindent
In this section, we experimentally examine our method, comparing its performance with state-of-the-art methods. We consider the standard benchmarks of TextVQA~\cite{singh2019towards}, ST-VQA\footnote{We use ST-VQA for denoting the dataset proposed in~\cite{biten2019scene}, and STVQA for denoting the general task of scene text VQA.)}~\cite{biten2019scene} and OCR-VQA~\cite{mishra2019ocr}. For pre-training we consider the same datasets used in~\cite{yang2021tap} with the addition of the Industrial Document Library (IDL)\footnote{\url{https://www.industrydocuments.ucsf.edu/}}. The IDL is a collection of industry documents hosted by UCSF. It hosts millions of documents publicly disclosed from various industries like tobacco, drug, food etc. The data from the website is crawled, which produced about 13M documents, translating to about 64M pages of various document images. IDL has various documents (like forms, tables, letters) with varied layouts. We further extracted OCR for each document using Textract OCR\footnote{\url{https://aws.amazon.com/textract/}}. Implementation details and further information on all datasets can be found in Appendix~\ref{appendix:implementation} and \ref{appendix:dataset}, respectively. We note that throughout the rest of the paper, $\ddagger$ refers to the models fine-tuned with both TextVQA and ST-VQA, at the same time. ``-Base'' and ``-Large'' model sizes refer to architectures that have 12+12 and 24+24 layers of transformers in encoder and decoder, respectively.

\paragraph{TextVQA Results}
Similar to previous work~\cite{yang2021tap}, we define two evaluation settings. The former is the constrained setting that only uses TextVQA for training and Rosetta for OCR detection. The latter is the unconstrained setting, in which we present our best performance with the state-of-the-art.
The first part of \cref{table:textvqa} reports the accuracy on TextVQA under the constrained setting. As can be appreciated, \AlgoName achieves the same performance as TAP~\cite{yang2021tap} without any pre-training, demonstrating the effectiveness of our model. 
Furthermore,  when \AlgoName is pre-trained on IDL, performance increase from 44.06\% to 48.38\% (\textbf{+4.32\%}) using the Rosetta OCR. This clearly shows the effectiveness of \textit{layout-aware} pre-training on scanned documents to the task of scene text VQA, even in the constrained setting. 

In the bottom part of \cref{table:textvqa} we modify the OCR system to a more recent one than Rosetta and gradually add additional training datasets (unconstrained settings). In this work, we experiment with Amazon Text-in-Image (\OCRNameNoSpace)\footnote{{\fontsize{7}{8}\selectfont \url{https://docs.aws.amazon.com/rekognition/index.html}}}~\cite{ughetta2021old}. As seen, when using \OCRName our method outperforms the M4C baseline, improving performance from 47.84\% to 52.29\% (\textbf{+4.45\%}). Furthermore, when enabling pre-training, \AlgoName outperforms the previous art \cite{yang2021tap} by large margins from 54.71\% to 58.03\% (\textbf{+3.32\%}) on validation and from 53.97\% to 58.86\% (\textbf{+4.89\%}) on the test. We note that for \cite{yang2021tap} there is a -0.74\% decrease between validation and test while for \AlgoName we observe an increase of +0.83\%, demonstrating better generalization.
Another critical point is that \AlgoName can benefit more when ST-VQA dataset is added as an extra fine-tune data. We believe this point to be critical since we do not have to train separate models for TextVQA and ST-VQA but rather one model that can get the best performance on both dataset. Finally, increasing our model capacity to \AlgoNameNoSpace-Large further boosts performance to 61.6\% (\textbf{+7.6\%} from \cite{yang2021tap}).

\begin{figure*}[t]
  \centering
  \includegraphics[width=0.95\textwidth]{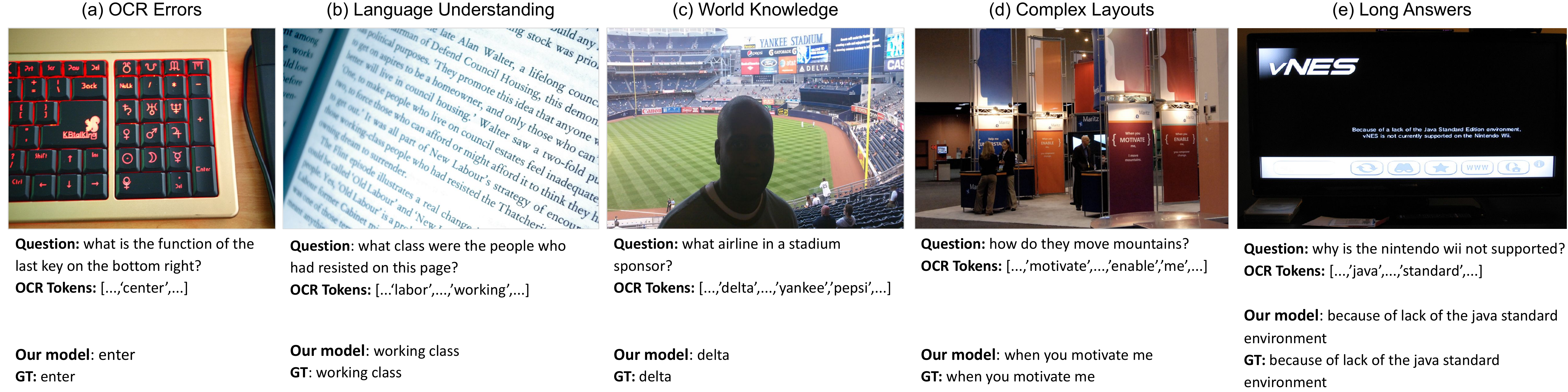}
  \caption{\textbf{Why is \TaskName hard?} Current state-of-the-art methods struggle to acquire various abilities which are needed for scene text VQA. We depict five representative abilities; fixing OCR errors, language understating, world knowledge, understating complex layouts and the ability to produces long answers. Our model is able to correctly answer each one of the these examples. We refer the reader to more qualitative results and comparisons to previous art in \cref{appendix:qualitative_examples}.}
  \label{fig:qualitative_fig}
  \vspace{-0.45cm}
\end{figure*}

\paragraph{ST-VQA Results}

\begin{table}
\normalsize
\begin{center}
\small
\bgroup
\def\arraystretch{1.1}
        \begin{tabular}{lccc}
        \toprule
        \textbf{Method} & \textbf{Val Acc.} & \textbf{Val ANLS} & \textbf{Test ANLS}\\
        \midrule
        M4C~\cite{hu2020iterative} & 38.05 & 0.472 & 0.462 \\
        SA-M4C~\cite{kant2020spatially} & 42.23 & 0.512 & 0.504 \\
        SMA~\cite{gao2020structured} & - & - & 0.466 \\
        CRN~\cite{liu2020cascade} & - & - & 0.483 \\
        LaAP-Net~\cite{han2020finding} & 39.74 & 0.497 & 0.485 \\
        LOGOS~\cite{lu2021localize} & 48.63 & 0.581 & 0.579 \\
        TAP~\cite{yang2021tap} & 50.83 & 0.598 & 0.597 \\
        \rowcolor{LightGray}
        \AlgoNameNoSpace-Base & 58.41 & 0.675 & 0.668 \\
        \rowcolor{LightGray}
        \AlgoNameCombinedNoSpace-Base & 59.09 & 0.683 & 0.684 \\
        \rowcolor{LightGray}
        \AlgoNameCombinedNoSpace-Large & \textbf{61.64} & \textbf{0.702}  & \textbf{0.696} \\
        \bottomrule
    \end{tabular}
\egroup
\tiny
\caption{\textbf{Results on the ST-VQA Dataset~\cite{biten2019scene}}. Our model advances the state-of-the-art performance by +10.81\%.}
\label{table:stvqa}
\vspace{-0.7cm}
\end{center}
\end{table}

\cref{table:stvqa} presents the accuracy on ST-VQA~\cite{biten2019scene} in the unconstrained setting. \AlgoName uses the \OCRName and is pre-trained on IDL and fine-tuned on the training set of ST-VQA. \AlgoNameCombined is also fine-tuned with TextVQA. The behaviour observed in TextVQA is consistent with ST-VQA dataset, \AlgoNameCombinedNoSpace-Base and \AlgoNameCombinedNoSpace-Large outperforming the previous art \cite{yang2021tap} by \textbf{+8.26\%} and \textbf{+10.81\%}, respectively. Moreover, we show a similar trend on OCR-VQA~\cite{mishra2019ocr} dataset where the discussion and the numbers can be found in \cref{sec:ocrvqa}. 

\paragraph{Qualitative Analysis}
In \cref{fig:qualitative_fig} we depict five different question categories which are representative of the capabilities \TaskName models need. We start with the ability to correct OCR errors (\cref{fig:qualitative_fig}~(a)). Most state-of-the-art OCR systems for scene text~\cite{baek2019wrong,litman2020scatter,fang2021read,nuriel2021textadain} operate on a word-level, and thus are unable to utilize image-level context. Current \TaskName methods depend on a pointer network for decoding, which means they are bounded by the performance of the OCR system at hand. Contrary to that, \AlgoName leverages image-level context and jointly with its generative nature, is able to correct OCR errors. Next, scene text VQA models are required to have the ability to understand language together with world knowledge (\cref{fig:qualitative_fig}~(b)(c)). Both requirements are met in \AlgoName thanks to its extensive pre-training.

As seen in \cref{fig:qualitative_fig}~(d), answering questions often requires reasoning over the relative spatial positions of the text in the image. Over the years several methods aimed at developing spatially aware models were proposed~\cite{kant2020spatially,lu2021localize}. However, most of those methods are complex, not easy to implement and eventually led to minimal performance improvements. \AlgoName is pre-trained on documents with layout information, which leads to a spatially aware model without any complex architectural changes. The last category we analyze is long answers (\cref{fig:qualitative_fig}~(e)). In practice, the existing pointer network decoding mechanism is also limited in ability to produce long answers. Furthermore, when pre-training is done on natural images as in \cite{yang2021tap}, the model hardly encounters long sentences. \AlgoName does not rely on a pointer network and is pre-trained on documents, in which text appears in a variety of lengths.

We provide further qualitative analysis and comparisons to previous work~\cite{hu2020iterative} in \cref{appendix:qualitative_examples}. In addition, we display failure cases of our method on the TextVQA dataset. The failure cases are mostly composed of OCR errors, compositionality of spatial reasoning and visual attributes. 


\begin{table}
\normalsize
\begin{center}
\small
\bgroup
\def\arraystretch{1.1}
        \begin{tabular}{lll}
        \toprule
        \textbf{Model}  & \textbf{OCR} & 
        \textbf{Acc.}\\
        \midrule
        T5-Base & Rosetta-en & 16.05 \\
        T5-Base & \OCRNameNoSpace & 21.93 \\
        T5-Base & GT text & 25.45 \\
        \bottomrule
    \end{tabular}
\egroup
\tiny
\caption{
\textbf{Zero Shot Performance of T5 Language Model on TextVQA.} In this setting, T5-Base is pre-trained on C4 and fine-tuned on SQuAD~\cite{rajpurkar2016squad}, a reading  comprehension dataset. Showing that a ``blind'' pre-trained language model can get up to 25.45\%.}
\label{tab:zero-shot}
\vspace{-0.7cm}
\end{center}
\end{table}

\section{Ablation Studies}
In this section, we provide insightful experiments which we deem crucial for the \TaskName task and its future development. We start off by showing the significance of language understanding in \TaskName.
Then, we show the effectiveness of language and layout information and discuss the biases existing in \TaskName benchmarks. Next, we study the effect of pre-training as a function of dataset size and type. Finally, we showcase our model's robustness towards vocabulary and OCR errors. All the numbers are obtained by using the TextVQA validation set.

\paragraph{Zero-shot Language Models on TextVQA}
To quantify the importance of language understanding in \TaskName, we devise a novel zero-shot setting where we use the T5 language model pre-trained on C4 and only fine-tuned on SQuAD~\cite{rajpurkar2016squad}, a reading comprehension dataset.
\cref{tab:zero-shot} presents the performance of this setting while varying the OCR system. Interestingly, even without any visual features or fine-tuning, T5 reaches a performance of 16.05\% and 21.93\% with Rosetta and \OCRNameNoSpace, respectively. 
More importantly, a zero-shot ``blind'' model with the perfect OCR (ground truth OCR annotation) can get to as high as 25\%, experimentally demonstrating the need for language understanding in \TaskName. However, one needs to be careful attributing the entirety of the performance to language understanding since deep models are known to exploit dataset biases~\cite{torralba2011unbiased}. Thus, we investigate if there are any biases in the data and if it is possible to categorize them.

\paragraph{Dataset Bias or Task Definition?}
To get a better sense of the biases in TextVQA,
we start by training a model where only questions are given as input. 
As can be seen in \cref{tab:ablation}, our model is able to achieve 11.18\% in a task that requires reading and reasoning about the text without \textit{the text}. Next, we study the effect of the OCR system by dividing the information provided by it into text token transcription, reading order and 2-D positional information. Reading order is the order where OCR tokens are extracted from left to right and top to bottom with respect to line boxes or text blocks.
Reading order is so intertwined with OCR systems that it is not thought of as a detached feature.


\begin{table}
\normalsize
\begin{center}
\small
\bgroup
\def\arraystretch{1.1}
        \begin{tabular}{cccccc}
        \toprule
        \textbf{Model} & \textbf{2-D} &\textbf{Pre-training} & \textbf{OCR} &  \textbf{Visual} & \textbf{ Acc.} \\ 
        \midrule
        \multirow{8}{*}{\AlgoName} & \ding{55} &\ding{55}  &\ding{55} & \ding{55} & 11.18 \\
        &\ding{55} & \ding{55}  & \ding{55} & $V$ & 11.74 \\
        & \ding{55} & \ding{55} & \textit{random} & \ding{55} & 41.77 \\
        \cdashline{2-6}
        & \ding{55} & \ding{55} & \checkmark & \ding{55} & 50.37 \\
        & \checkmark &\ding{55}  & \checkmark & \ding{55} & 51.22\\
        
         &\checkmark & \ding{55}  & \checkmark & $V$ & 52.29 \\
        \cdashline{2-6}
         & \checkmark &\checkmark  & \checkmark & \ding{55} & 57.38 \\
         &\checkmark & \checkmark  & \checkmark & $F$ & \textbf{58.11} \\
         & \checkmark &\checkmark  & \checkmark & $V$ & 58.03 \\
         \hline
         \multirow{3}{*}{\AlgoNameCombined} &\checkmark  & \checkmark  & \checkmark & \ding{55}&  58.92\\
       & \checkmark  &\checkmark  & \checkmark & $F$ & 58.45\\
         & \checkmark  & \checkmark  &\checkmark & $V$ & \textbf{59.53}\\
						
        \bottomrule
    \end{tabular}
\egroup
\tiny
\caption{\textbf{\AlgoName Ablation Studies on TextVQA.} We ablate \AlgoName-Base by varying the building blocks of our method, including pre-training, input types and fine-tuning data. $V$ refers to ViT and $F$ refers to FRCNN as visual backbone, \textit{random} means OCR tokens are provided but presented in a random reading order.}
\label{tab:ablation}
\vspace{-0.7cm}
\end{center}
\end{table}

As shown in \cref{tab:ablation}, adding OCR tokens without any reading order gives us 41.77\% and a fixed reading order already gets us to 50.37\%, showing the importance of reading order for given OCR tokens.
The gain becomes marginal when adding the 2-D positional and visual information without pre-training, +0.85\% and +1.09\%, respectively. However, when performing \textit{layout-aware} pre-training on documents, obtaining alignment between the layout information and the semantic representations, \AlgoNameNoSpace's performance increases significantly by +7.01\% to 57.38\%.
In other words, we can already achieve SOTA on a \textit{Visual} Question Answering task without any visual features (other than using the images for OCR extraction). Finally, adding visual features still \textit{marginally} increases performance by around +0.7\%. Recently, \cite{wang2021towards} showed a similar phenomenon using the M4C~\cite{hu2020iterative} architecture, where visual information only slightly contributed to the performance, validating that this is not specific to our technique.



\begin{table}
\normalsize
\begin{center}
\small
\bgroup
\def\arraystretch{1.1}
        \begin{tabular}{lcl}
        \toprule
        \textbf{Model} & \textbf{Pre-training Data} & \textbf{Acc.}\\
        \midrule
        \multirow{6}{*}{\AlgoNameNoSpace-Base} & \ding{55} &  50.37  \\
        & TextVQA & 51.81 \\
         & \scriptsize{TextVQA,ST-VQA,TextCaps,OCR-CC} & 54.22 \\
         & IDL - 1M & 55.12    \\
         & IDL - 11M &  56.28  \\
         & IDL - 64M & 58.03   \\
         & \scriptsize{IDL-64M,TextVQA,ST-VQA,TextCaps,OCR-CC} &  \textbf{58.51}  \\
        \cdashline{1-3}
        \multirow{2}{*}{\AlgoNameCombinedNoSpace-Base} & IDL - 64M & \textbf{59.53}   \\
        & \scriptsize{IDL-64M,TextVQA,ST-VQA,TextCaps,OCR-CC} & 59.06 \\
        \bottomrule
    \end{tabular}
\egroup
\tiny
\caption{\textbf{The Effect of Pre-training.} Ablation studies on pre-training as a function of different datasets type and size.}
\label{tab:pretraining}
\vspace{-0.7cm}
\end{center}
\end{table}

Regarding the comparison of the different visual backbones, we train our model with visual features extracted either from FRCNN~\cite{anderson2018bottom} or ViT~\cite{dosovitskiy2020image}. We note that the performance difference is very marginal when only TextVQA is used in fine-tuning. However, when TextVQA and ST-VQA are used together, the model with FRCNN features perform worse than the model without any visual features while ViT increases performance by +0.61\%, demonstrating that ViT features can scale better with more data. 
 
At this point, we would like to take a step back and discuss \TaskName as a task. As we see it, our analysis can be interpreted from two viewpoints. The first viewpoint is how \TaskName is defined as a task. In particular, is the \TaskName task defined such that all (or a majority of) questions should require reasoning over all modalities (including visual features)? Regardless of the answer, we present a second viewpoint, a dataset bias. To better explore the bias perspective, in \cref{appendix:dataset_bias} we visualize question-image pairs sorted by the information required to answer them. Clearly, generating questions from the final category (\ie questions which require reasoning over all modalities) is not an easy task. Furthermore, we quantitatively showed that at-least 60\% of the questions do not fall under the final category, allowing the model to extensively exploit language priors and make educative guesses.
Both viewpoints lead us to wonder are visual features even needed for \TaskNameNoSpace? Or better yet, is vision an artifact in \TaskName task? We believe that visual features are of importance for the task of \TaskNameNoSpace, however current benchmarks do not reflect it, making it harder to evaluate how much V matters in \TaskNameNoSpace.

\paragraph{The Effect of Large-Scale Pre-Training}

\cref{tab:pretraining} demonstrates the benefits of pre-training while varying the datasets type and scale.
First, we explore the effect of pre-training on natural images with visual features (as done in \cite{yang2021tap}) using our architecture. In particular, we add the image-text matching objective and leverage the same datasets (which we term TAP-datasets) as in~\cite{yang2021tap}. Pre-training only on TextVQA (\cref{tab:pretraining}), provides only +1.5\% improvement for us compared to~\cite{yang2021tap} reporting +5\%. The same behaviour of diminished gain is also observed with TAP-datasets.

Next, we compare IDL and TAP-datasets in pre-training. Even pre-training on 1M documents, \AlgoNameNoSpace's performance increases by almost +5\%, which is more than the combination of all TAP-datasets. This is inspiring for two reasons, one of which is 1M documents are less than two thirds the size of TAP-datasets~\cite{yang2021tap}. Secondly, our model is pre-trained with a simple de-noising objective and no visual features, making the pre-training significantly faster (around 23 times) compared to TAP~\cite{yang2021tap} which is pre-trained with visual features, scene text features and multiple losses. We also argue that IDL is a better bed for \textit{layout-aware} pre-training since it provides varied layouts to better align with language.
Finally, we discuss the effect of increasing the size of IDL.
Adding an order of magnitude more data only result in +1\% or +2\% increase. We emphasize that 64M documents hardly seems the saturation point for \AlgoNameNoSpace, \ie more pre-training data can still improve the performance, especially when also increasing the model capacity.

\paragraph{Vocabulary Reliance and Robustness Towards OCR Errors}
\begin{table}
\normalsize
\begin{center}
\small
\bgroup
\def\arraystretch{1.1}
        \begin{tabular}{lcccc}
        \toprule
        \multirow{2}{*}{\textbf{Model}} & \textbf{All} & \textbf{InVoc.} & \textbf{OutVoc.} & \textbf{Gap} \\
        & 5000 & 3731 & 1269 & \\
        \midrule
        M4C \cite{hu2020iterative} & 47.84 &  51.07 & 38.37 & \color{Red}{\textbf{12.7}} \\
        \AlgoNameNoSpace-Base & 59.53 &  59.93 & 58.35 & \color{Red}{1.58} \\
        \bottomrule
    \end{tabular}
\egroup
\tiny
\caption{\textbf{Vocabulary Reliance.} Accuracy gap between answers with words in and out of vocabulary used by~\cite{hu2020iterative,yang2021tap,kant2020spatially}. InVoc. and OutVoc. stand for in and outside the vocabulary, respectively.}
\label{tab:vocab_dependency}
\vspace{-0.7cm}
\end{center}
\end{table}
Current state-of-the-art methods predict the answer through an amalgamation of a pointer mechanism and a dataset-specific 5K most frequent vocabulary. The usage of a vocabulary is limiting in a real-world scenario and may result in high performance on in-vocabulary answers but lead to poor performance on out-of-vocabulary ones, in other words, lack of generalization. This is clearly observed in \cref{tab:vocab_dependency} where M4C~\cite{hu2020iterative} exhibits a heavy reliance on the fixed vocabulary as the gap between categories is \textbf{-12.7\%}.
Contrary to that, \AlgoName is not limited to any handcrafted dataset-specific vocabulary. Its gap between in and out of the training vocabulary is only \textbf{-1.58\%}.

\begin{figure}[t]
    \centering
    \includegraphics[width=0.9\columnwidth]{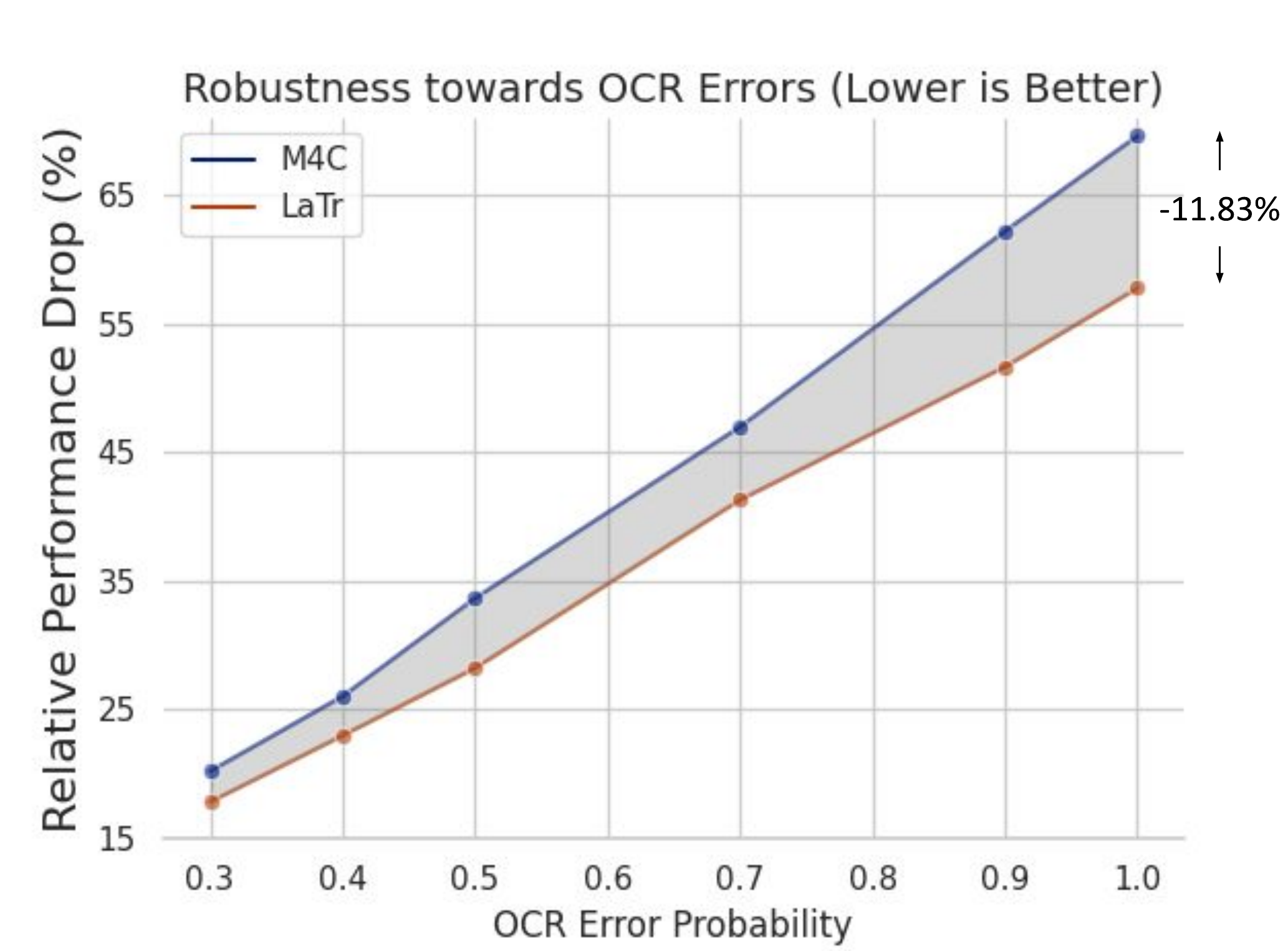}
    \caption{\textbf{Robustness towards OCR Errors.} OCR Error Probability refers to the percentage of OCR tokens that we replace a single character by a random one, simulating OCR engine errors. \AlgoNameNoSpace's relative robustness is higher compared to \cite{hu2020iterative} and increases with the probability of OCR errors.}
    \label{fig:ocr_robust}
    \vspace{-0.5cm}
\end{figure}

Finally, we experimentally display that our model is more robust to OCR errors compared to M4C architecture. To validate our claim we introduce a new setting where we replace a single character for certain amount of OCR tokens. Whether to replace a character in each word is decided according to the threshold from a Bernoulli distribution, called OCR Error Probability in \cref{fig:ocr_robust}. 
To simulate real-world OCR errors, we utilized the publicly available nlp-augmenter from \cite{ma2019nlpaug}. \AlgoName is more robust than \cite{hu2020iterative} and in fact the lead increases as more OCR errors are added.

\section{Conclusion}

We convey a couple of important take-home messages for the \TaskName community. Firstly, \textit{language and layout are essential}. Language indirectly is utilized for questions that need world/prior knowledge or simply for language understanding. Layout information allows the model to reason over spatial relations. In our work, we methodologically demonstrated their importance to \TaskNameNoSpace. Secondly, we propose a \textit{layout-aware} pre-training and show a new symbiosis between scanned documents and scene text, where the layout information of scanned documents promotes a better understanding of scene text information. This is exciting news since scanned documents are more abundantly available than natural images that contains scene text.
Text in documents appears in a variety of complex layouts, making our model spatially aware without any complex architectural changes. Last but not least, we replace the extensive need of FRCNN for feature extraction. We exhibit that using a ViT as a feature extractor can scale better than FRCNN, \ie leading to better performance. However, perhaps more crucially, we diagnose a condition in which \TaskName models (ours included) make use of the visual features \textit{marginally}. 
This begs the question whether this is because of the dataset bias, and we as a community need to make V matter again in VQA.

{\small
\bibliographystyle{ieee_fullname}
\bibliography{egbib}
}

\clearpage

\appendix
\section{Implementation Details}
\label{appendix:implementation}
In this section we detail the implementation specifics of our paper divided into three parts; (1) pre-training; (2) fine-tuning; (3) ablation studies. In our work all models are pre-trained on 8 A100 GPUs and are implemented using PyTorch~\cite{paszke2017automatic}. T5 uses SentencePiece~\cite{kudo2018sentencepiece} to encode the text as WordPiece tokens~\cite{kudo2018subword, sennrich2015neural}, we use a vocabulary of 32,000 wordpieces for all experiments.

\noindent
\paragraph{Pre-training.} For the base-size model, we utilize a batch size of 25 for each GPU with the maximum OCR token length set to 512 and pre-training is done for 2.2M steps. For the large-size model, we use a batch size of 28 for each GPU with the maximum OCR token length set to 384 and pre-training is done for 0.9M steps. In both models, the learning rate is increased linearly over the warm-up period of 100K steps to 1e-4 learning rate and then linearly decayed to 0 at the end of the training, and we enable gradient accumulation. For our \textit{layout-aware} de-noising task, we corrupt 15\% of the original text sequence, with a span length which vary as a function of the amount of text in each sample. 

\noindent
\paragraph{Fine-tuning.} We train all of our models for 100K steps and use AdamW~\cite{loshchilov2017decoupled} optimizer with 1e-4 max learning rate. Warm-up period is set to 1,000 steps and again is linearly decayed to zero. The same batch sizes that were used for pre-training are also used in this stage. We use a ViT~\cite{dosovitskiy2020image} to extract visual features. The ViT is pre-trained and fine-tuned on ImageNet~\cite{deng2009imagenet} for classification. We follow the implementation and use the weights from HuggingFace \cite{wolf2019huggingface}~\footnote{\url{https://huggingface.co/transformers/model\_doc/vit.html}}.
\noindent

\paragraph{Ablation studies.} For ablating the visual backbone, we follow the common practice \cite{hu2020iterative, singh2019strings, singh2019towards, yang2021tap} of detecting objects with a Faster R-CNN detector~\cite{anderson2016spice} which is pre-trained on the Visual Genome dataset~\cite{krishna2017visual}. We keep the 100 top-scoring objects per image and, similarly to previous work, only fine-tune the last layer. We now detail the specifics of our pre-training ablation studies. When exploring the effect of pre-training with visual features, we combine the de-noising pre-training task with an image-text (contrastive) matching (ITM) task. For the ITM taks, we follow the same implementation as in \cite{yang2021tap}, the text input is polluted 50\% of the time by replacing the whole text sequence with a randomly-selected one from another batch. The polluted text words are thus not paired with the visual patch features from the ViT. The ITM task takes the sequence feature as the input and aims to predict if the sequence has been polluted or not. One important point to mention is that for the de-noising task, we compute the gradients for both encoder and decoder. Yet, for the ITM task, we merely compute the gradients for our encoder. 

For the vocabulary reliance experiment, we collect the top 5000 frequent words from the answers
in the training set as our answer vocabulary as done by \cite{hu2020iterative, yang2021tap}.

\section{Datasets}
\label{appendix:dataset}
\label{appendix:idl}
\begin{figure}[t]
    \centering
    \includegraphics[width=\columnwidth]{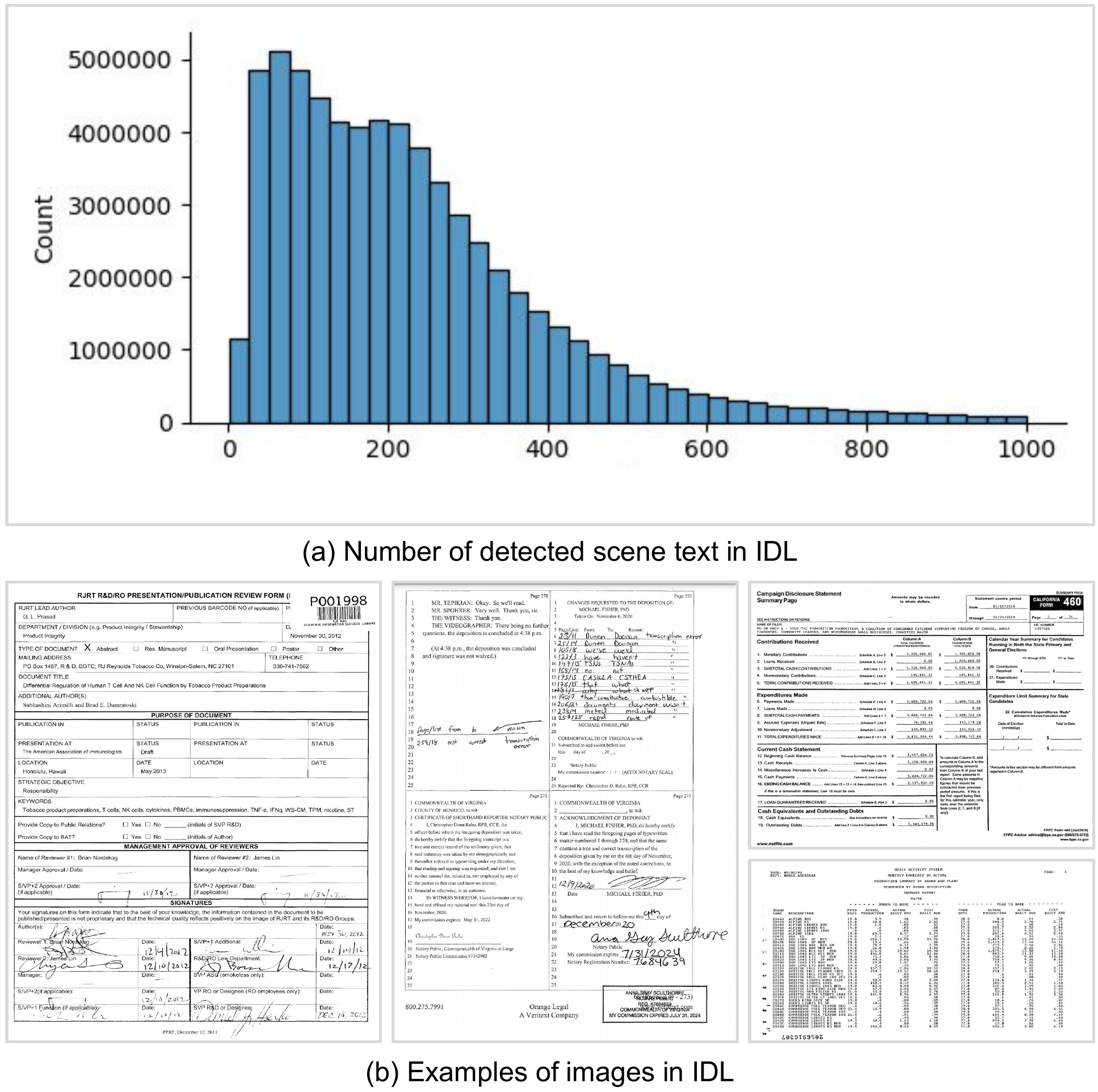}
    \caption{\textbf{IDL dataset.} (a) We show the distribution of the detected OCR number by Textract OCR~\cite{litman2020scatter,aberdam2021sequence,nuriel2021textadain} on the IDL dataset. (b) We visualize representative examples from the dataset.}
    \label{fig:idl70m_word_freq_plot}
\end{figure}

\noindent
\textbf{TextVQA~\cite{singh2019towards}} contains 28k images from the Open Images~\cite{kuznetsova2020open} dataset. The questions and answers are collected through Amazon Mechanical Turk (AMT) where the workers are instructed to come up with questions that require reasoning about the scene text in the image. Following VQAv2~\cite{goyal2017making}, 10 answers were collected for each question. In total, there are 45k questions divided into 34,602, 5,000, and 5,734 for train, validation and test set, respectively.

\noindent
\textbf{ST-VQA~\cite{biten2019scene}} is an amalgamation of well-known computer vision datasets, namely: ICDAR~2013\cite{karatzas2013icdar}, ICDAR2015~\cite{karatzas2015icdar}, ImageNet~\cite{deng2009imagenet}, VizWiz~\cite{gurari2018vizwiz}, IIIT Scene Text Retrieval~\cite{MishraICCV13_IIIT_text}, Visual Genome~\cite{krishna2017visual} and COCO-Text~\cite{veit2016coco}. 
ST-VQA is also collected through AMT, asking workers to come up with questions so that the answer is always the scene text in the image. In total, there are 31k questions, separated into 26k questions for training and 5k questions for testing.

\noindent
\textbf{TextCaps~\cite{sidorov2020textcaps}} is composed of 28,408 images, when there are 5 captions per image, amounting to a total of 142,040 captions. The images are taken from TextVQA~\cite{singh2019towards} dataset. The dataset is annotated with AMT. The AMT annotators are asked to provide captions that are based on the text in the image. In other words, the captions can not be generated without having OCR tokens, however, the provided captions do not necessarily contain the OCR tokens.

\noindent
\textbf{OCR-VQA~\cite{mishra2019ocr}} is composed of 207,572 images of book covers and contains more than 1 million question-answer pairs about these images. The questions are template-based, asking about information on the book such as title, author, year. The questions are all can be answered by inferring the book cover images. 

\noindent
\textbf{OCR-CC~\cite{yang2021tap}} is a subset of Conceptual Captions (CC)~\cite{sharma2018conceptual} dataset proposed by \cite{yang2021tap}. This subset is compromised of 1.367 million scene text-related image-caption pairs. To obtain OCR-CC, \cite{yang2021tap} used the Microsoft Azure OCR system to extract the text in the image, then any image that does not contain any text or any image that only has watermarks is discard. As this subset is not publicly released, we follow the same process to create it. However, we use Amazon-OCR\footnote{{\fontsize{7}{8}\selectfont \url{https://docs.aws.amazon.com/rekognition/index.html}}} as our main OCR system. 
As was presented in \cite{yang2021tap}, the distribution of the detected scene text in the original CC datasets is that only 45.16\% of the images contain text. Out of the images that do contain text, the data has a mean and median of 11.4 and 6 scene text detected per image.

\section{The Industrial Document Library dataset}

In this subsection, we present more details on the Industrial Document Library (IDL)\footnote{\url{https://www.industrydocuments.ucsf.edu/}} dataset. As mentioned in the main paper, the IDL is a digital archive of documents created by industries which influence public health. The IDL is hosted by the University of California, San Francisco Library. It hosts millions of documents publicly disclosed from various industries like drug, chemical, food and fossil fuel. The data from the website is crawled, leading to about 13M documents, which translate to about 70M pages (64M usable) of various document images. IDL has various documents (like forms, tables, letters) with varied layouts as seen in \cref{fig:idl70m_word_freq_plot}~(b). We extracted OCR for each document using Textract OCR\footnote{\url{ttps://aws.amazon.com/textract/}}~\cite{ughetta2021old}.

The crawled and OCR'ed IDL data was pre-processed before consuming for pre-training. We removed all documents which had less than 10 words or the image was unreadable. In addition, to weed out documents having a majority of erroneous OCR text and documents with non-English content, we considered a fixed English dictionary with a 350K-sized vocabulary and check if each OCR word is part of that dictionary with either exact-match or edit-distance of 1. We do not apply this filter if the word is either a number, float, currency or date (as those are unlikely to be present in the fixed English dictionary and would inflate the error count if considered). If the number of erroneous words are $\geq$ 50\% for that document, we ignore it. 
After all this filtering we are left with about 64M documents (roughly 6M are discarded) which are used for pre-training. The subsets used in \cref{tab:pretraining} are uniform random samples of this larger 64M data.

We show in \cref{fig:idl70m_word_freq_plot}~(a) the detected OCR word distribution across all the 64M documents. The plot roughly looks like a right-skewed normal distribution, with the majority of documents lying in the hump (having 20 to 400 words per doc). Unlike OCR-CC, documents by definition contain words, and thus we are able to use over 91\% of the original IDL dataset (compared to 45.16\% for OCR-CC). In addition, as clearly seen, there are much more words an average in IDL than OCR-CC which is extremely beneficial for pre-training in scene text VQA tasks. In \cref{fig:idl70m_word_freq_plot}~(b) we depict representative examples from the IDL dataset.

\section{OCR-VQA Results}
\label{sec:ocrvqa}
    

\begin{table}
\normalsize
\begin{center}
\small
\bgroup
\def\arraystretch{1.1}
        \begin{tabular}{lcc}
        \toprule
        \textbf{Method} & \textbf{Val Acc.} & \textbf{Test Acc.}\\
        \midrule
        CNN~\cite{mishra2019ocr} & -& 14.3 \\
        BLOCK~\cite{mishra2019ocr} & -& 42.0 \\ 
        BLOCK+CNN~\cite{mishra2019ocr} & -& 41.5 \\
        BLOCK+CNN+W2V~\cite{mishra2019ocr} & -& 48.3 \\
        M4C~\cite{hu2020iterative} & 63.5 & 63.9 \\
        
        \rowcolor{LightGray}\AlgoNameNoSpace-Base & \textbf{67.5} & \textbf{67.9} \\
        \bottomrule
    \end{tabular}
\egroup
\tiny
\caption{\textbf{Results on the OCR-VQA Dataset~\cite{mishra2019ocr}}. We use our base model pretrained on IDL and utilize Rosetta OCR system so that it is comparable across all the models. \AlgoName improves the state-of-the-art by +4.0\%.}
\label{table:ocrvqa}
\vspace{-0.9cm}
\end{center}
\end{table}
As commonly done by previous work~\cite{hu2020iterative}, we only evaluate our model using the constrained setting. In this setting, we do not change the OCR system, \ie we use Rosetta OCR system. Similarly to TextVQA and ST-VQA datasets, \AlgoNameNoSpace-Base outperforms the previous state of the art \cite{hu2020iterative} by a large margin, specifically, from 63.5\% to 67.5\% (\textbf{+4.0\%}).

\section{Qualitative Examples}
\label{appendix:qualitative_examples}
\begin{figure*}[t]
    \makebox[\linewidth]{
    \includegraphics[height=0.92\textheight]{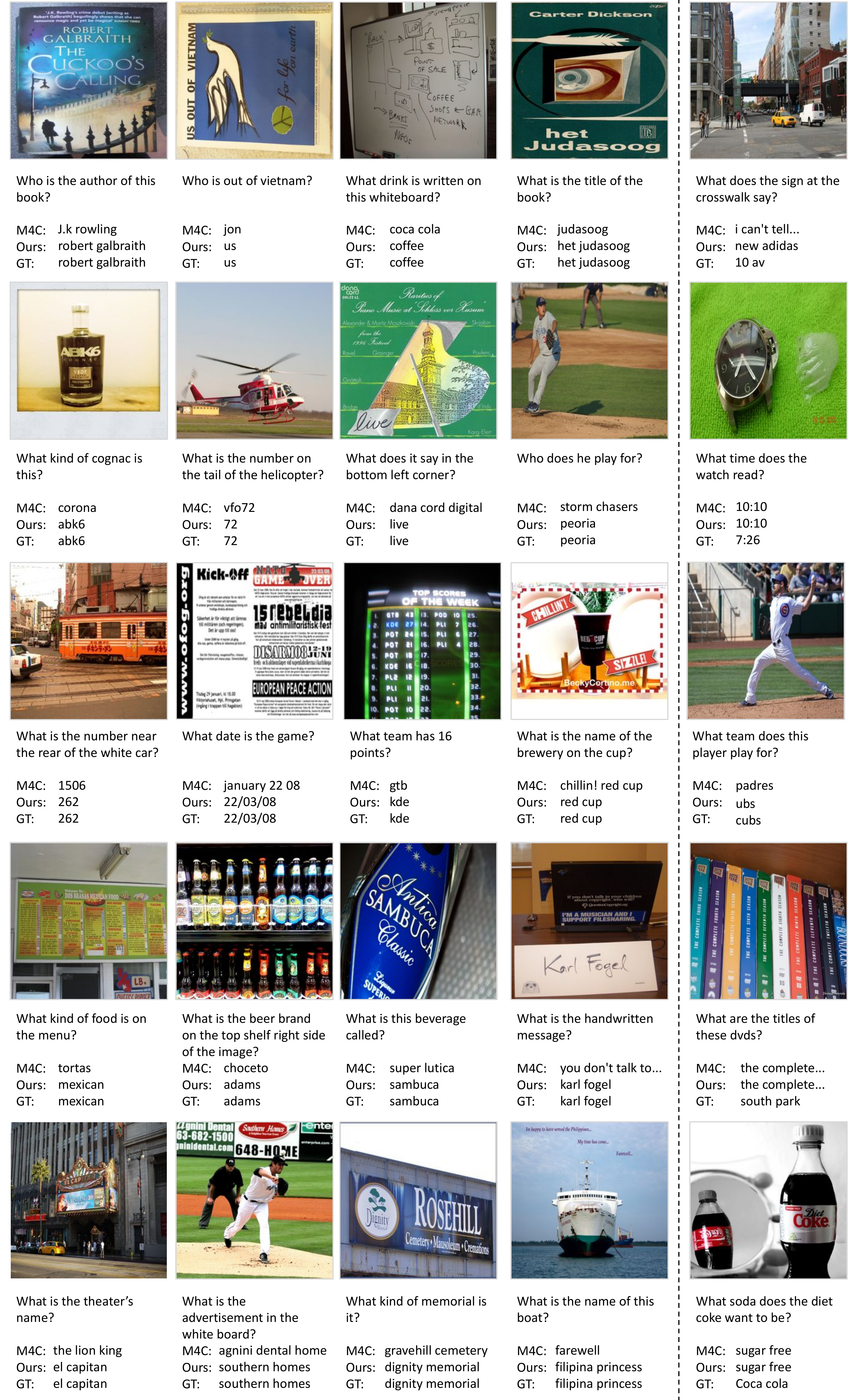}
    }
    \caption{\textbf{Qualitative Examples.} The first four columns displays failure cases of M4C~\cite{hu2020iterative} in which our model is successful. As can be seen, \AlgoName is able to outperform M4C on a variety of different question types, including, layout, world knowledge, natural language understand and more. In the last column, we present fail cases of our model, demonstrating representative failure cases of \AlgoNameNoSpace. We note that we present the questions as they are originally appear in the TextVQA dataset~\cite{singh2019towards}}
    \label{fig:qualitative_examples}
\end{figure*}

In this section, we present additional qualitative examples of our method compared with M4C~\cite{hu2020iterative}. In the first four columns of \cref{fig:qualitative_examples}, we display examples in which our model is successful while M4C fails. Compared to M4C, our model clearly has better natural language understanding (top left image). In addition, our model has the ability to reason over layout information significantly better than M4C (third image in row 3). This is both attributed to the extensive pre-training and the fact that we leveraged documents for performing \textit{layout-aware} pre-training with 2-D spatial position embedding.

Out of the cases displayed, we wish to further discuss two types of observed biases in the data. The first is for the question asking ``\textit{what is the handwritten message?}''. Our model successfully answers this question, both with and without visual features. This indicates that, at-least for the model without the visual features, the model is just guessing based on some heuristic. In this case, it could be that the largest OCR bounding box is the most probable answer. As all the datasets were created by AMT it is possible that the annotators created most of the questions base on the largest or the clearest text in the image. The second type of observed bias is the fact that most images contain only a few pieces of text. Thus, the model can make a lot of educated guesses. For example, the question asking ``\textit{What is the number on the rear of the white car?}''. There are only two numbers in the image, thus giving the model at-least 50\% chance of guessing correctly. Similarly, more than 85\% ``Yes/No'' questions are with answers “Yes" in TextVQA dataset, given the model a strong (and incorrect) prior knowledge, allowing easy guesses.

An additional interesting observation is with regard to questions about reading the time from an analog watch. We observed that both our model and the M4C model, in most cases, predict the time of 10:10 regardless of the actual time in the image. This is a bias the models developed from a common marketing trick. Watch sellers displays watches aimed to 10:10 as business marketing research showed it increases sales, and therefore, our model can't actually read the time but just guesses the most likely time based on the pre-training prior knowledge.

In the final column of \cref{fig:qualitative_examples}, we display our model's failure cases. The failure cases are mostly composed of OCR errors, compositionality of spatial reasoning and visual attributes. We wish to further discuss the last example (bottom right) as we believe this is an example of a question which requires a higher level of ``intelligence'' than the other examples. To answer this question, the model has to not only reason over both the image and the text, but also to understand that the soda wish to be like the regular coca-cola as it is "imagining" its reflection in the mirror.

\section{Dataset Bias or Task Definition?}

\label{appendix:dataset_bias}
\begin{figure*}[t]
    \makebox[\linewidth]{
    \includegraphics[height=0.9\textheight]{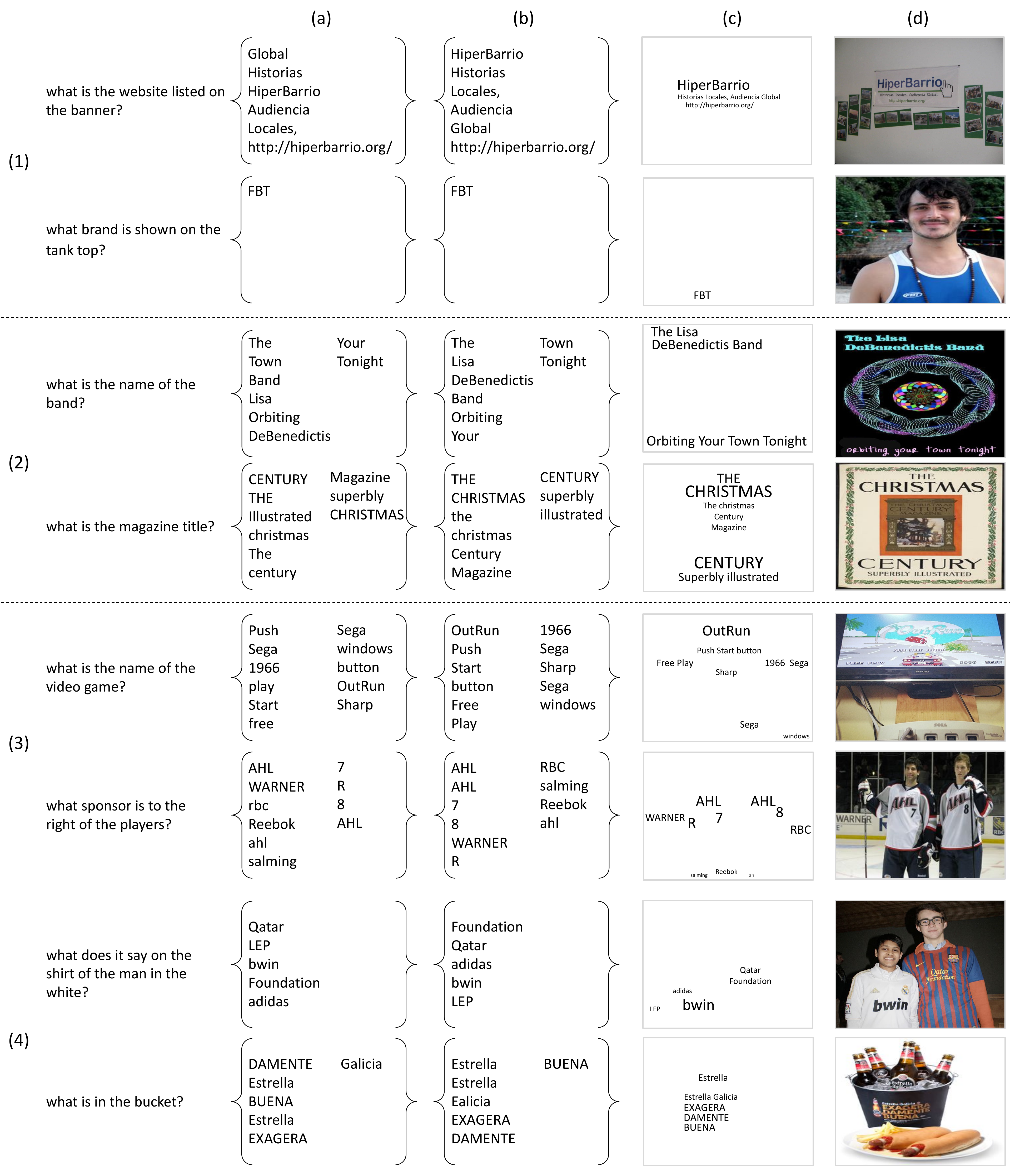}
    }
    \caption{\textbf{Dataset Bias or Task Definition?.} We depict four different questions types based on the information needed to answer them. Questions which require; (a) order-less bag-of-words; (b) ordered bag-of-words; (c) words and their 2-D spatial layout; (d) words, their 2-D spatial layout and the image.}
    \label{fig:dataset_bias}
\end{figure*}

In the main paper, we showed that STVQA models (ours included) make use of the visual features marginally. This begs the question whether this is because of a dataset bias, or is it simply the task nature. To explore this, we attempt to categorize the type of questions current benchmarks consist of. We divide the questions in TextVQA~\cite{singh2019towards} into four different categories. The questions categories are defined by the information type required to answer them. The first category consist of all questions that can be answered with just an order-less bag of words. In \cref{fig:dataset_bias}~(1) we depict examples from this category, \ie question that do not require anything beyond the order-less bag-of-words and some world knowledge. Base on the analysis presented in the main paper, this category amounts to over 40\% of the test data and include the questions that can be answered with just the questions ($\approx$11\%). The second category consists of questions which require an ordered bag-of-words. Currently, most papers treat the OCR system as a black-box and reading order is so intertwined with OCR systems that it is not thought of as a detached feature. We make the distinction between the information types extracted from the OCR system and demonstrate that an additional 10\% of the questions can be answered by just adding the reading order. Examples from this category are depicted in \cref{fig:dataset_bias}~(2).

The next category requires to reason over both word tokens and their 2-D spatial layout. In the main paper, we showed that via \textit{layout-aware} pre-training, we are able to leverage the additional layout information to boost performance by over 7\%. Base on a qualitative analysis, we believe that 7\% is the lower bound of this category size and more questions can be answered by just reasoning over the text and its layout. Examples from this category can be found in \cref{fig:dataset_bias}~(3). 
The last category consists of question which require reasoning over all modalities, specifically the text, the layout information and the image itself. Generating such questions is not an easy task, and therefore in current benchmarks most question do not fall under this category. We believe that in order to advance the field of STVQA this issue needs to be addressed.
We propose a simple mechanism for determining whether an image falls under the last category. In this mechanism the question is given to the annotator with just the words and layout visualization (third column of \cref{fig:dataset_bias}), if the question can still be answered it should be dropped. Examples from this category are depicted in \cref{fig:dataset_bias}~(4).


\end{document}